\documentclass[10pt, a4paper]{article}

\usepackage[final]{lrec2026} 
\usepackage[most]{tcolorbox}
\usepackage{tabularx}
\usepackage{ragged2e}
\usepackage{enumitem}
\usepackage{booktabs} 
\usepackage{multicol}
\usepackage{multirow}

\usepackage{footmisc}

\usepackage{listings}
\usepackage[most]{tcolorbox}
\newtcblisting[auto counter]{codelisting}[2][]{sharp corners, 
    fonttitle=\footnotesize\bfseries, colframe=gray, listing only, 
    listing options={basicstyle=\scriptsize\ttfamily, breaklines=true, breakautoindent=false, breakindent=0pt, escapeinside={(*@}{@*)}
}, 
    title= #2, #1}

\newcommand{\phiIV}{phi-4 (14B) }
\newcommand{\phiIVdef}{phi-4 (14B)}
\newcommand{\qwenIV}{Qwen3 (4B) }
\newcommand{\qwenXIV}{Qwen3 (14B) }
\newcommand{\qwendef}{Qwen3 (4B,14B)}
\newcommand{\olmo}{Olmo-3 (7B) }
\newcommand{\olmodef}{Olmo-3-7B-Instruct}
\newcommand{\llama}{Llama-3.1 (8B) }
\newcommand{\llamadef}{Llama-3.1-8B-Instruct}

\newcommand{\ournamedataset}{EarlySciRev}
\title{\ournamedataset: A Dataset of Early-Stage Scientific Revisions Extracted from LaTeX Writing Traces}

\name{Léane Jourdan\textsuperscript{1}, Julien Aubert-Béduchaud\textsuperscript{1}, Yannis Chupin\textsuperscript{1}, \\ {\bf \large Marah Baccari\textsuperscript{1} and Florian Boudin\textsuperscript{2}}}

\address{\textsuperscript{1}Nantes Université, École Centrale Nantes, CNRS, LS2N, UMR 6004, F-44000 Nantes, France\\ \textsuperscript{2}Inria, LS2N, Nantes Université, France \\
         \{firstname.lastname\}@univ-nantes.fr\\}

\abstract{
Scientific writing is an iterative process that generates rich revision traces, yet publicly available resources typically expose only final or near-final versions of papers. 
This limits empirical study of revision behaviour and evaluation of large language models (LLMs) for scientific writing.
We introduce \ournamedataset, a dataset of early-stage scientific text revisions automatically extracted from arXiv LaTeX source files. 
Our key observation is that commented-out text in LaTeX often preserves discarded or alternative formulations written by the authors themselves.
By aligning commented segments with nearby final text, we extract paragraph-level candidate revision pairs and apply LLM-based filtering to retain genuine revisions.
Starting from 1.28M candidate pairs, our pipeline yields 578k validated revision pairs, grounded in authentic early drafting traces. 
We additionally provide a human-annotated benchmark for revision detection.
\ournamedataset{} complements existing resources focused on late-stage revisions or synthetic rewrites and supports research on scientific writing dynamics, revision modelling, and LLM-assisted editing.
 \\ \newline  }

\begin{document}

\maketitleabstract

\section{Introduction}

Academic writing is an inherently demanding task, requiring precision, clarity, and conciseness, often under time pressure and frequently in a second language.
To support this process, researchers increasingly rely on LLMs, which can rewrite entire passages in response to high-level instructions. 
These models offer unprecedented assistance for revising drafts, improving readability, and articulating complex ideas with clarity and fluency.

However, assessing the impact of LLMs on scientific writing remains challenging. 
%
The key data required for such analysis, \emph{drafts, revisions, and writing traces}, are largely inaccessible.
%
While most scientific papers are publicly available in their final or near-final form, the iterative writing process that produced them typically remains hidden.

Because scientific writing is inherently incremental, authors progressively refine arguments, restructure paragraphs, clarify explanations, and adjust claims.
%
This process naturally generates various intermediate artifacts, including discarded sentences, rewritten paragraphs, and commented alternatives.
%
Yet these traces are rarely preserved in publicly accessible resources. As a result, existing research on revision modelling predominantly relies on late-stage revisions (e.g., between submission versions) or synthetic rewrites. Early drafting stages, where substantial conceptual and structural changes occur, remain underexplored.

The lack of access to early-stage revisions limits our ability to study authentic writing dynamics, measure quality improvements, train models that support in-depth revision, and evaluate the role of LLMs in scientific writing. 
It also hinders systematic investigation of issues such as stylistic homogenisation, bias propagation, or factual distortion introduced during automated rewriting. 

%
%
%
This inaccessibility largely stems from ethical, legal, and ownership concerns: drafts may contain personal information, unpublished ideas, or confidential comments, and they often relate to papers that later fall under publisher copyright.

In this paper, we introduce \ournamedataset, a large-scale dataset of early-stage scientific revisions automatically extracted from arXiv LaTeX source files.
Our key insight is that LaTeX comments often preserve discarded or intermediate versions of sentences and paragraphs.
%
By mining these commented segments and aligning them with nearby final text, we recover fine-grained revision pairs that reflect authentic author rewriting.
We present a complete pipeline\footnote{\url{https://github.com/JourdanL/EarlySciRev}} for (i) collecting computer science papers from arXiv, (ii) cleaning and processing LaTeX sources, (iii) extracting candidate revision pairs from commented text, and (iv) filtering genuine revisions using a LLM-based classification.
Finally, we report an annotation study that benchmarks several LLMs on the revision detection task, and use the best model to curate the final dataset.


Our contributions are threefold:
\begin{itemize}[topsep=.2em, itemsep=.1em]
    \item We propose a new method to retrieve early-stage scientific writing traces by exploiting commented content in LaTeX source files.
    \item We introduce \ournamedataset{}, a large-scale dataset of paragraph-level scientific revisions automatically extracted from arXiv papers, capturing authentic early drafting revisions.\footnote{\url{https://huggingface.co/datasets/taln-ls2n/EarlySciRev}\label{refnote}}
    \item We release a human-annotated benchmark for revision detection in scientific text, enabling systematic evaluation of both LLMs and future revision models.\footref{refnote}
\end{itemize}


\section{Related Work}

A variety of datasets for text revision have been released over the years, reflecting growing interest in modelling writing and rewriting processes.
Some of these datasets rely on synthetic revisions, generated either automatically~\citep{ito-etal-2019-diamonds} or manually by annotators who are not the original authors~\citep{mita-etal-2024-towards}.
While such resources are useful for controlled experimentation, they do not capture authentic authorial revision behaviour.

Among datasets that feature real author revisions, two primary sources have emerged: arXiv~\citep{tan-lee-2014-corpus,du-etal-2022-read,jiang-etal-2022-arxivedits} and OpenReview~\citep{darcy2023aries,jourdan-etal-2024-casimir,jourdan-etal-2025-pararev}.
These resources enable the study of revision behaviour in real-world scientific writing contexts.

Despite their usefulness, existing datasets have clear limitations.
%
First, many are restricted in scope.
Some focus exclusively on abstracts~\citep{tan-lee-2014-corpus,du-etal-2022-read}, while others remain relatively small in scale, ranging from a few hundred~\citep{jiang-etal-2022-arxivedits,mita-etal-2024-towards,ruan-etal-2024-re3} to a few thousand papers~\citep{kuznetsov-etal-2022-revise,darcy2023aries,dycke-etal-2023-nlpeer,lin-moprd-2023,jourdan-etal-2025-pararev}.

Second, most existing datasets primarily capture late-stage revisions, typically between near-final drafts posted on arXiv or submission platforms.
%
These revisions often reflect already polished manuscripts prepared for public dissemination, rather than the exploratory and formative stages of writing.
Early drafting phases, where substantial conceptual, structural, and stylistic changes are made, are therefore largely absent from current resources.
To our knowledge, the only dataset that explicitly targets writing traces from the earliest stages of the drafting process is ScholaWrite~\citep{wang2025scholawritedatasetendtoendscholarly}. 
%
However, it is limited to only five papers, which restricts its applicability beyond exploratory analysis.



As a result, access to early-stage writing traces remains a major bottleneck for studying authentic scientific revision behaviour and for developing models that support in-depth revision.
%
%
In this work, we hypothesize that such traces can be recovered directly from the LaTeX source files uploaded to arXiv. 
In practice, authors frequently leave commented-out sentences, paragraphs, or alternative phrasings in the source code (i.e.~lines beginning with the ``\texttt{\%}'' character). 
Mining these commented segments offers a promising and underexplored avenue for reconstructing fine-grained, early-stage scientific revisions.

\section{Data Creation}
\label{sec:data-creation}

This section describes the pipeline used to construct the dataset, from collecting raw data from arXiv to cleaning LaTeX sources and extracting candidate revision pairs.

\subsection{Data Collection}

We rely on the \texttt{arxiv-metadata-oai-snapshot.json}\footnote{\url{https://www.kaggle.com/datasets/Cornell-University/arxiv}} dump downloaded on January 21, 2026. 
This metadata file contains 2,932,928 arXiv repositories, among which 596,118 are distributed under a permissive licence.\footnote{\href{http://creativecommons.org/licenses/by-nc-sa/4.0/}{CC BY-NC-SA 4.0}, \href{http://creativecommons.org/licenses/by-sa/4.0/}{ CC BY-SA 4.0}, \href{http://creativecommons.org/licenses/by/4.0/}{CC BY 4.0}, \href{http://creativecommons.org/licenses/publicdomain/}{Public Domain List}, \href{http://creativecommons.org/licenses/by-nc-sa/3.0/}{CC BY-NC-SA 3.0}, \href{http://creativecommons.org/licenses/by/3.0/}{CC BY 3.0}, \href{http://creativecommons.org/publicdomain/zero/1.0/}{CC0 1.0}}



In this work, we focus on computer science (CS) papers, which represent 286,747 with a valid licence.
For each of these papers, we downloaded all available source archives (zip files), resulting in approximately 1.2 TB of source data.

\subsection{LaTeX Source Processing}

We first filter the source files to retain only LaTeX documents that contain potentially meaningful comments. 
Specifically, we select files that include commented lines that do \emph{not} start with a backslash. 
This criterion excludes commented-out LaTeX commands, while preserving comments that may contain previous versions of sentences or paragraphs.

We then apply the following cleaning steps:
\begin{enumerate}[topsep=.3em, itemsep=.1em]
    \item We retain only the content between \texttt{\textbackslash begin\{document\}} and \texttt{\textbackslash end\{document\}}.
    \item We remove non-textual environments, including \texttt{table}, \texttt{figure}, \texttt{align}, \texttt{tikz}, and \texttt{algorithm}.
    \item We replace each \texttt{equation} environment with a special token \texttt{[EQUATION]}, as equations may appear within running text.
    \item We remove LaTeX commands that do not contain textual content (e.g., \texttt{\textbackslash appendix}, \texttt{\textbackslash vspace}), while preserving structural commands such as \texttt{\textbackslash section}.
\end{enumerate}

These steps ensure that the resulting text primarily consists of natural language content suitable for revision analysis.

    
\subsection{Candidate Revision Pair Extraction}

Our objective is to extract candidate revision pairs composed of:
i) a block of uncommented text that appears in the compiled document (the \textit{final paragraph}), and 
(ii) a block of commented text that may correspond to an earlier version (the \textit{commented paragraph}).
%
%
We define a \textit{block} as a sequence of consecutive lines of the same type (commented or uncommented), not interrupted by an empty line or by a line of the other type.



For each commented block $c$, we compare it to the five preceding and five following blocks. 
For each neighbouring block that corresponds to a final paragraph $f$, we compute a normalised difference ratio based on the Levenshtein distance between the two texts.
To account for cases in which a revision affects only part of a paragraph, we compute this distance using a sliding window over the final paragraph.
We define this normalised difference ratio as:
\begin{equation}
d_{norm}(f, c) = lev(f, c) / \max(|f|, |c|)
\end{equation}
where $lev(\cdot,\cdot)$ denotes the Levenshtein distance and $|\cdot|$ the character length.
We treat a pair as a candidate revision when $d_{norm} < 0.7$.
This threshold was set empirically on a subset of the dataset.


For each final paragraph, we retain all associated candidate commented revisions.
Before concatenating uncommented blocks, we remove inline trailing comments to ensure that the resulting text matches the content of the compiled document.

\section{Data Annotation}

The automatic extraction procedure described above result in 1,269,976 candidate revision pairs. 
However, not all of these candidates correspond to genuine rewriting instances.
We therefore introduce an additional filtering step based on LLMs. 
To select an appropriate prompting strategy and model, we first construct a human-annotated gold subset.

\subsection{Human Annotation}

The objective of the annotation campaign is to determine whether a given pair of paragraphs constitutes a genuine revision instance.
We randomly sampled 500 candidate pairs, each consisting of a \textit{commented paragraph} (representing a potential original version) and a corresponding \textit{final paragraph}. 
The annotation was carried out by five annotators: two master's students, two junior researchers, and one senior researcher. 
All annotators had prior experience with scientific writing and NLP research. 

For each paragraph pair, annotators answered the following binary question: \emph{“Can the final paragraph be qualified as a revision of the original one(s)?”}
%
Annotators were provided with detailed guidelines specifying decision criteria (see Appendix~\ref{annex:guidelines}).
These guidelines were derived from the revision taxonomy proposed by~\citet{jourdan-etal-2025-pararev} and adapted to the present task.

Annotation was conducted using \textit{Label Studio}. Paragraph pairs were displayed side by side. 
%
To facilitate comparison, identical text spans shared by the two paragraphs were  highlighted (Appendix~\ref{annex:examples}).

\begin{figure}[t]
    \centering
    \includegraphics[width=.9\linewidth]{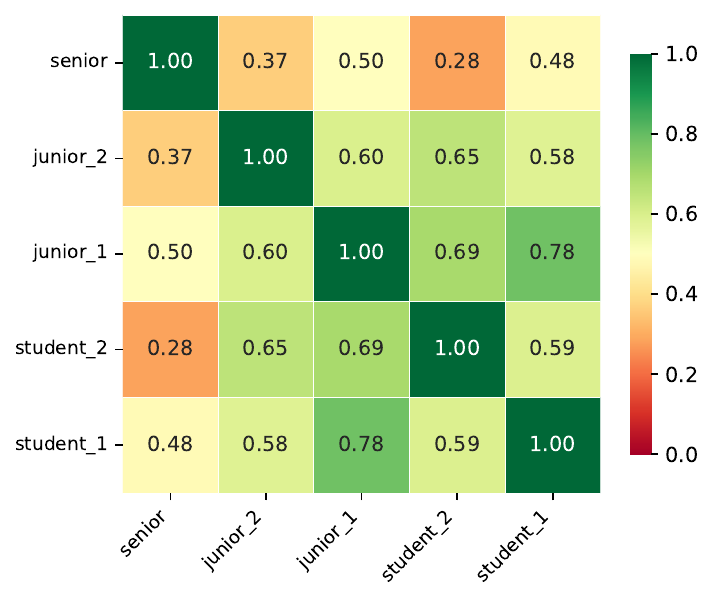}
    \caption{Pairwise Cohen's Kappa ($\kappa_{\text{Cohen}}$) scores between annotators.} 
    \label{fig:CohenAgreement}
\end{figure}

\textbf{Inter-annotator agreement.}
Figure~\ref{fig:CohenAgreement} shows pairwise Cohen's $\kappa_{\text{Cohen}}$ scores between annotators.
Agreement varies across annotator pairs, with lower agreement observed for pairs involving the senior annotator.The senior annotator was more selective than other annotators on the pairs considered as revision.
%

To assess overall inter-annotator agreement, we apply Fleiss' $\kappa_{\text{Fleiss}}$ under a partially overlapping design, in which each item is annotated by three annotators selected from the pool of five.
The resulting $\kappa_{\text{Fleiss}} = 0.54$ indicates moderate agreement, according to the scale proposed by~\citet{landis1977measurement}.


\subsection{LLM-Based Filtering}

Given the number of the extracted candidate pairs, manual annotation of the full dataset is infeasible.
%
We therefore rely on automatic classification to identify genuine revision pairs.
To select the most reliable approach, we evaluate several models and prompting strategies on the human-annotated subset.
%
Since each item in this subset was labelled by three annotators, we use majority vote to derive the reference label for evaluation.

\textbf{Approaches.} We compare two LLM-based approaches: i) \textbf{Standard prompting}, where the model directly answers the binary question posed in the prompt, and the decision is extracted from the generated output. ii) \textbf{*-PLUIE}~\citep{lemesle2026pluie}, a perplexity-based LLM-as-a-judge method that estimates the model's confidence without generating free-form text.
%
In the *-PLUIE setting, the model estimates its confidence by computing the perplexity of candidate responses (``Yes'' vs. ``No'') given the prompt.
Positive values favour the revision hypothesis, negative values the opposite, and a binary decision is obtained by thresholding the score (default threshold 0, optimised a posteriori when annotations are available).

\begin{figure}
    \centering
    \input{prompt/llm_prompt.tex}
    \caption{Prompt used in both *-PLUIE and LLM-choice settings.}
    \label{fig:prompt_llm}
\end{figure}

\textbf{Models and setup.}
We evaluate several LLMs of varying sizes, including \qwendef, \phiIVdef, \olmodef, and \llamadef. 
All models are run in a chat-completion setting using \texttt{bfloat16} precision.
Sampling is disabled during inference (\texttt{do\_sample=False}) to ensure deterministic outputs.
%
%
%
The same prompt (Figure~\ref{fig:prompt_llm}) is used across both standard prompting and *-PLUIE configurations, enabling a direct comparison between generation-based and perplexity-based classification for each model.

\begin{table}[]
    \centering
    \resizebox{\linewidth}{!}{

\begin{tabular}{cl | cccc}
    \toprule
    \multicolumn{2}{c}{\textbf{Model}}  & \textbf{Acc.} & \textbf{P.} &\textbf{R.} &  \textbf{time}\\
    \midrule
    \parbox[t]{2mm}{\multirow{10}{*}{\rotatebox[origin=c]{90}{{\small \textbf{*-PLUIE}} }}}

    &\qwenIV (thr=0)  & 0.77 & 0.77 & 0.76 & \multirow{2}{*}{22h}\\
    &\qwenIV (thr=-4.75) & 0.78 & 0.75 & 0.81 & \\ 
    \cmidrule(rl){2-6}
    
    &\olmo (thr=0) &  0.74 & 0.73 & 0.76& \multirow{2}{*}{35h}\\
    &\olmo (thr=0.60) & 0.74 & 0.78 & 0.67 &\\ 
    \cmidrule(rl){2-6}

    &\llama (thr=0) &  0.70 & 0.65 & 0.85 & \multirow{2}{*}{35h}\\
    &\llama (thr=0.85) & 0.73 & 0.75 & 0.67 &\\ 
     \cmidrule(rl){2-6}

    &\phiIV (thr=0) &  0.77 & 0.71 & 0.92 & \multirow{2}{*}{55h}\\
    &\phiIV (thr=2.15) & 0.79 & 0.75 & 0.85 &\\ 
    \cmidrule(rl){2-6}
    
    &\qwenXIV (thr=0) & \underline{0.80} & 0.75 & 0.90 & \multirow{2}{*}{62h}\\
    &\qwenXIV (thr=5.55)& \textbf{0.82} & \underline{0.80} & 0.86 &\\

    \midrule
    \parbox[t]{2mm}{\multirow{5}{*}{\rotatebox[origin=c]{90}{{\small \textbf{LLM-choice}} }}}
    &\qwenIV  & 0.59 & 0.55 & \underline{0.95 }& 29h\\
    &\llama   & 0.59 & 0.55 & \textbf{0.97} & 46h\\
     &\olmo    & 0.68 & \textbf{0.81} & 0.45 & 45h \\
     &\phiIV   & 0.78 & \underline{0.80} & 0.75 & 81h \\
    &\qwenXIV  & \underline{0.80} & 0.78 & 0.83 & 82h\\
    \bottomrule
  \end{tabular} 
    }
    
    \caption{Alignment of LLM-based classifier to human majority vote. \textit{Thr.} is the threshold used to binarise *-PLUIE values, \textit{Acc.} the accuracy, \textit{P.} the precision, \textit{R.} the recall and \textit{time} the estimated time to classify all the data. \textbf{Bold} values indicate the best results, and \underline{underlined} values indicate the second-best results.}
    \label{tab:results_llm}
\end{table}

\textbf{Results and model selection.}
Results are reported in Table~\ref{tab:results_llm}. 
Across both approaches, \qwenXIV achieves the best overall performance, with accuracy exceeding 80\%.
Balancing classification performance and computational efficiency, we select the *-PLUIE configuration with \qwenXIV to filter the full dataset, using the optimal threshold determined on the annotated subset.


\section{Dataset Statistics}

\begin{table}[]
    \centering
    \resizebox{\linewidth}{!}{
    
 \begin{tabular}{ccccc}
    \toprule
    \textbf{\#rev} & \textbf{\#paper} & \textbf{\#rev/\S} & \textbf{\#words/\S} &\textbf{\% words diff}\\
    \midrule
    578,440 & 104,023 & 1.10 & 82.42 & 56.85 \\
    \bottomrule
  \end{tabular} 

    }
    
    \caption{Characteristics of \ournamedataset. In this order: number of revision pairs, number of articles, average number of commented paragraphs per final paragraph, average number of words per paragraph (final version), average percentage of difference in words per revision pair}
    \label{tab:stats}
\end{table}

Applying the selected LLM-based filtering strategy retain 578,440 revision pairs out of the initial 1,2M candidates (approximately 45.55\%).
%
These validated revisions correspond to 523,932 distinct final paragraphs. Among them, 46,192 paragraphs are associated with more than one revision candidate, reflecting cases where authors experimented with multiple alternative formulations before settling on a final version or cases where multiple previous paragraphs were merged into one.
At the document level, the revisions are distributed across 104,023 articles.
Those characteristics and more are summarised in Table~\ref{tab:stats}.

Qualitative inspection suggests that revisions captured in EarlySciRev range from local fluency edits to more substantial restructuring and clarification of scientific arguments, to draft ideas left as to do with their fully written version.

The 500 paragraphs used for human annotation are also included and filtered at this step. Both the human annotated dataset  and large LLM filtered one are openly available.\footnote{\url{https://huggingface.co/datasets/taln-ls2n/EarlySciRev}}
\section{Conclusion}
\label{sec:conclusion}

We introduced \ournamedataset, a dataset of paragraph-level scientific revisions extracted from LaTeX writing traces in arXiv source files, together with a human-annotated benchmark for revision detection.
By focusing on early drafting stages, this resource makes visible revision phenomena that are typically inaccessible in existing datasets.

%
%

%
%
\ournamedataset{} supports empirical study of scientific writing dynamics and provides a foundation for developing and evaluating revision models, including LLM-based systems.
To facilitate reproducibility and further development, we release the full extraction and filtering framework, enabling updates as new papers become available.\footnote{\url{https://github.com/JourdanL/EarlySciRev}}

A future step could be to label all data with a revision intention and see how the distribution compare to dataset focusing on late stage revision.

\section{Limitations}
The current pipeline is restricted to computer science papers, as we only process licensed CS articles from arXiv. Writing practices and revision behaviours may differ across disciplines, limiting the generalizability of our findings. Extending the approach to other domains is a natural direction for future work.

Additionally, we do not explicitly control for the language of the paper. A part of papers submitted to arXiv are written in languages other than English. Our LLM-based filtering relies on prompts written in English, which may affect classification reliability for non-English texts. Adapting prompts to the language of each document or incorporating language identification into the pipeline could improve robustness.

\section*{Acknowledgements}
This work was partly supported the AID-CNRS NaviTerm project (convention 2022 65 0079 CNRS Occitanie Ouest).

\section{Bibliographical References}\label{sec:reference}

\bibliographystyle{lrec2026-natbib}
\bibliography{custom.bib}


\appendix

\onecolumn
\section{Annotation Guidelines}

\begin{tcolorbox}[
    breakable,              
    colback=white,
    colframe=black,
    boxrule=0.8pt,
    arc=0mm,
    width=\linewidth,
    fontupper=\scriptsize,  
    left=4pt, right=4pt, top=4pt, bottom=4pt,
    before skip=10pt,
    after skip=10pt,
    title={Annotation Guidelines for Revision Detection}, 
    fonttitle=\bfseries\small, 
    colbacktitle=gray!10, 
    coltitle=black
]

\vspace{0.5em}
\textbf{1. Introduction}

The goal of this annotation campaign is to detect text revisions amid paragraphs originating from computer science scientific papers. A paragraph level revision is defined as a paragraph that is substantially modified for clarity, simplicity, style and other aspects. To that end, some final paragraphs have been selected and each one of them was provided with one or more original paragraphs that were under comment in the latex file. A final paragraph is a paragraph that is not commented and is suspected to be a revision of an original paragraph(s). 
In this task, we aim to characterize the final paragraphs’ relationship with the suspected original paragraph(s), so that they can be classified as revisions or not down the line.

\vspace{0.5em}
\textbf{2. Annotation Task}

Annotators are presented with a pair of paragraphs: an original version composed of one or several paragraphs and a final version. Their task is to answer the following question: \textit{Can the final paragraph be qualified as a revision of the original one(s)?}

Annotators must select one of the following labels:
\begin{itemize}[noitemsep, topsep=0pt, leftmargin=*]
    \item \textbf{YES}: The final paragraph constitutes a revision of the original paragraph.
    \item \textbf{NO}: The final paragraph does not constitute a revision (e.g., different scientific content, the idea developed is not the same, introduces too much new information, or does not change the text).
\end{itemize}
As several original candidates are proposed, the annotator can answer Yes for multiple paragraphs (e.g. in cases of paragraph merging or iterative revision).

\vspace{0.5em}
\textbf{2.1 Positive example}

\noindent
\begin{tabularx}{\linewidth}{|X|X|}
\hline
\textbf{Original Paragraph} & \textbf{Final paragraph} \\
\hline
Therefore, the generalization rapidly decreases after augmentationinterrupted when training with a single background because the learning direction toward generalization about various backgrounds is not helpful to train. On the other hand, the training can have helpwhen their difculty is solved by augmentation, such as Figure 2(b) and Figure 2(c). & 
Therefore, the generalization rapidly decreases after augmentation is interrupted during training with a single background because the learning direction toward generalization about various backgrounds is not helpful to train. In contrast, the training can help when their difficulty is solved by augmentation (Figure 2(b), 2(c)). \\
\hline
\end{tabularx}

\vspace{0.5em}
\textbf{2.2 Negative example}

\noindent
\begin{tabularx}{\linewidth}{|X|X|}
\hline
\textbf{Original Paragraph} & \textbf{Final paragraph} \\
\hline
In future research, the multi-mode characteristics will be studied to improve the representativeness of degradation features and the trendability of HI, and transfer learning approaches will be investigated to improve the generalization ability of the proposed framework and extend it to different systems. & 
Based on the ablation study, it can be concluded that the proposed SkipAE, inner HI-prediction block, and the HI-generating module jointly improve the ability of HI for reliable and accurate prognostics. \\
\hline
\end{tabularx}

\vspace{0.5em}
\textbf{2.3 Annotation Procedure}\\
For each pair of paragraphs (original and final), annotators must proceed as follows:
\begin{enumerate}[noitemsep, topsep=0pt, leftmargin=*]
    \item Read the final paragraph carefully to understand its scientific content and intent.
    \item Read the original paragraph to identify any differences with respect to the final version.
    \item Assess whether each original is rephrased in the final paragraph considering aspects such as: grammatical correctness, clarity and readability, fluency and coherence, appropriateness of scientific style.
    \item Determine whether the scientific meaning of the paragraph is preserved in the final version.
    \item Assign a label (YES or NO) according to the decision rules defined below.
\end{enumerate}
Annotators should base their decision solely on the information contained in the paragraph pair and should not rely on external context. Also, annotators are prohibited to invent things.

\vspace{0.5em}
\textbf{2.4 Decision Rules}\\
Annotators must apply the following rules when assigning labels:

\textbf{Assign YES} if at least one of the following conditions are met for a part or the whole paragraph:
\begin{enumerate}[noitemsep, topsep=0pt, leftmargin=*]
    \item The final paragraph is a revised version of the original paragraph, incorporating changes ranging from minor edits to substantial rephrasing.
    \item The final version has been modified through the addition, the substitution or the deletion of ideas or facts.
    \item The revision expands on the same idea with additional or withdrawn details.
    \item The differences between the original and final paragraphs indicate the correction of document processing errors (e.g., parsing issues, segmentation errors, or misaligned paragraphs).
\end{enumerate}

\textbf{Assign NO} if:
\begin{enumerate}[noitemsep, topsep=0pt, leftmargin=*]
    \item None of the above conditions are met.
    \item If the annotator is unsure whether the revised paragraph constitutes a valid paragraph-level revision.
    \item If there are only equations or code.
    \item If the two paragraphs are the exact same.
\end{enumerate}

If presented with multiple commented paragraphs for the same final paragraph, one or more commented paragraph can independently be considered as a revision. Classifying a commented paragraph as a revision does not disqualify the other proposed candidates. The same goes with the negative label : all the commented paragraphs may not qualify as a revision.

\end{tcolorbox}
\begin{center}
    \small \textbf{Table 3:} Detailed annotation guidelines provided to the annotators for the revision detection task.
    \label{annex:guidelines}
\end{center}

\section{Annotation Examples}

\begin{tabularx}{\textwidth}{|X|X|c|}
\hline
\textbf{Reference} & \textbf{Candidate} & \textbf{Revised} \\
\hline
\textcolor{green}{We} observed \textcolor{green}{that} \textcolor{green}{AR2VP} demonstrates \textcolor{green}{superior} \textcolor{green}{entity} \textcolor{green}{perception} outcomes, achieving the highest \textcolor{green}{overall} \textcolor{green}{perception} performance. \textcolor{green}{This} \textcolor{green}{analysis} underscores \textcolor{green}{that} current \textcolor{green}{V2X} \textcolor{green}{technologies} rarely rely on RSUs \textcolor{green}{to} expand \textcolor{green}{perception} \textcolor{green}{horizons.} \textcolor{green}{In} \textcolor{green}{contrast,} \textcolor{green}{AR2VP} harnesses \textcolor{green}{the} \textcolor{green}{latent} strengths \textcolor{green}{of} RSUs \textcolor{green}{to} address \textcolor{green}{intra-scene} \textcolor{green}{changes,} which enhances the vehicle's ability \textcolor{green}{to} adapt to dynamic scenes, consequently elevating the \textcolor{green}{overall} \textcolor{green}{perception} \textcolor{green}{capabilities.} \textcolor{green}{However,} \textcolor{green}{AR2VP} does exhibit \textcolor{green}{a} \textcolor{green}{performance} drawback \textcolor{green}{in} \textcolor{green}{pedestrian} \textcolor{green}{detection,} implying \textcolor{green}{a} particular challenge \textcolor{green}{in} \textcolor{green}{detecting} \textcolor{green}{small} \textcolor{green}{targets.} & \textcolor{green}{We} find \textcolor{green}{that} \textcolor{green}{AR2VP} shows \textcolor{green}{superior} \textcolor{green}{entity} \textcolor{green}{perception} results, with \textcolor{green}{overall} \textcolor{green}{perception} performance being the best. \textcolor{green}{This} \textcolor{green}{analysis} suggests \textcolor{green}{that} existing \textcolor{green}{V2X} \textcolor{green}{technologies} merely utilize RSU \textcolor{green}{to} extend \textcolor{green}{perception} \textcolor{green}{horizons.} \textcolor{green}{In} \textcolor{green}{contrast,} \textcolor{green}{AR2VP} leverages \textcolor{green}{the} \textcolor{green}{latent} advantages \textcolor{green}{of} RSU \textcolor{green}{to} model \textcolor{green}{intra-scene} \textcolor{green}{changes,} further enhancing vehicle adaptability \textcolor{green}{to} scene dynamics, thereby augmenting \textcolor{green}{overall} \textcolor{green}{perception} \textcolor{green}{capabilities.} \textcolor{green}{However,} in comparison to V2V, \textcolor{green}{AR2VP} exhibits \textcolor{green}{a} \textcolor{green}{performance} disadvantage \textcolor{green}{in} \textcolor{green}{pedestrian} \textcolor{green}{detection,} indicating \textcolor{green}{a} certain discrepancy \textcolor{green}{in} \textcolor{green}{detecting} \textcolor{green}{small} \textcolor{green}{targets.} & yes \\
\hline
\textcolor{green}{Under} \textbackslash cref\{ass:runtimecomplexity\} and following \textbackslash cref\{tab:setops\}, the outer approximative Minkowski sum from \textbackslash cref\{prop:minkSum\_polyzono\}, the Minkowski difference, and the linear map in \textcolor{green}{the} \textcolor{green}{computation} \textcolor{green}{of} \textcolor{green}{the} \textcolor{green}{outer} \textcolor{green}{approximation} \textcolor{green}{\$\textbackslash outerBRSAE\{-t\}\$} are all \$\textbackslash bigO\{n\textasciicircum 3\}\$, while the computation of the inner approximation \$\textbackslash innerBRSAE\{-t\}\$ \textcolor{green}{is} \textcolor{green}{dominated} \textcolor{green}{by} \textcolor{green}{the} conversion to a constrained zonotope, \textcolor{green}{which} \textcolor{green}{is} \$\textbackslash bigO\{n\textasciicircum 4\}\$. & \textcolor{green}{Under} \textbackslash cref\{ass:runtimecomplexity\}, \textcolor{green}{the} \textcolor{green}{computation} \textcolor{green}{of} \textcolor{green}{the} \textcolor{green}{outer} \textcolor{green}{approximation} \textcolor{green}{\$\textbackslash outerBRSAE\{-t\}\$} \textcolor{green}{is} marginally \textcolor{green}{dominated} \textcolor{green}{by} \textcolor{green}{the} over-approximative Minkowski sum, \textcolor{green}{which} \textcolor{green}{is} \$\textbackslash bigO\{(\textbackslash cons\{\}+2n)n, since the Minkowski difference and linear map are at most \$\textbackslash bigO\{(\textbackslash cons\{\}+2n)n \textbackslash steps\{\}n\}\$ and \$\textbackslash bigO\{(\textbackslash cons\{\}+2n)n\textasciicircum 2\}\$, respectively; all these operations are essentially \$\textbackslash bigO\{n\textasciicircum 3\}\$ for \$n \textbackslash gg \textbackslash steps\{\}\$ and under \textbackslash cref\{ass:runtimecomplexity\}. & yes \\
\hline
\textcolor{green}{\$\$\textbackslash sum\_\{t=1\}\textasciicircum T} \textcolor{green}{\textbackslash sum\_\{y} \textcolor{green}{\textbackslash in} \textcolor{green}{\textbackslash \{0,} \textcolor{green}{1\textbackslash \}\}} \textcolor{green}{p\_t\textasciicircum y} \textcolor{green}{\textbackslash hat\{\textbackslash ell\}\_t(y)} \textcolor{green}{-} \textbackslash inf\_\{j \textcolor{green}{\textbackslash in} [N]\}\textbackslash sum\_\{t=1\}\textasciicircum T \textbackslash hat\{\textbackslash ell\}\_t(\textbackslash mathcal\{E\}\_t\textasciicircum \{j\} ) \textcolor{green}{\textbackslash leq} \textcolor{green}{\textbackslash frac\{\textbackslash ln} \textcolor{green}{N\}\{\textbackslash eta\}} \textcolor{green}{+} \textcolor{green}{\textbackslash eta} \textcolor{green}{\textbackslash sum\_\{t=1\}\textasciicircum T} \textcolor{green}{\textbackslash hat\{\textbackslash ell\}\_t(1)} \textcolor{green}{+} \textcolor{green}{\textbackslash eta\textbackslash sum\_\{t=1\}\textasciicircum T} p\_t\textasciicircum 1(1 - p\_t\textasciicircum 1) \textbackslash hat\{\textbackslash ell\}\_t(0)\textasciicircum 2 + \textbackslash eta\textbackslash sum\_\{t=1\}\textasciicircum T p\_t\textasciicircum 1 \textbackslash hat\{\textbackslash ell\}\_t(1)\textasciicircum 2.\$\$ \textbackslash end\{lemma\} & \textcolor{green}{\$\$\textbackslash sum\_\{t=1\}\textasciicircum T} \textcolor{green}{\textbackslash sum\_\{y} \textcolor{green}{\textbackslash in} \textcolor{green}{\textbackslash \{0,} \textcolor{green}{1\textbackslash \}\}} \textcolor{green}{p\_t\textasciicircum y} \textcolor{green}{\textbackslash hat\{\textbackslash ell\}\_t(y)} \textcolor{green}{-} \textbackslash sum\_\{t=1\}\textasciicircum T \textbackslash sum\_\{y \textcolor{green}{\textbackslash in} \textbackslash \{0, 1\textbackslash \}\} \textbackslash hat\{\textbackslash ell\}\_t(\textbackslash mathcal\{E\}\_t\textasciicircum \{j\}) \textcolor{green}{\textbackslash leq} \textcolor{green}{\textbackslash frac\{\textbackslash ln} \textcolor{green}{N\}\{\textbackslash eta\}} \textcolor{green}{+} \textcolor{green}{\textbackslash eta} \textcolor{green}{\textbackslash sum\_\{t=1\}\textasciicircum T} \textcolor{green}{\textbackslash hat\{\textbackslash ell\}\_t(1)} \textcolor{green}{+} \textcolor{green}{\textbackslash eta\textbackslash sum\_\{t=1\}\textasciicircum T} \textbackslash sum\_\{y \textbackslash in \textbackslash \{0, 1\textbackslash \}\} p\_t\textasciicircum y \textbackslash hat\{\textbackslash ell\}\_t(y)\textasciicircum 2.\$\$ & no \\
\hline
\textbackslash iuhead\{Date \textcolor{green}{of} fault-triggering test creation and modification:\} We identified all \textcolor{green}{the} commits that are \textcolor{green}{associated} \textcolor{green}{with} \textcolor{green}{the} \textcolor{green}{fault-triggering} tests and analyzed when \textcolor{green}{the} commits happened (e.g., before or after the bug was reported/fixed). We used the git command ``\textbackslash inlineCode\{git log -L:[funcname]:[file]\}'' to identify the list of commits \textcolor{green}{that} modified the fault-triggering test and the modification date. & We collected the date and time information provided in the output \textcolor{green}{of} this command to track \textcolor{green}{the} development activities \textcolor{green}{associated} \textcolor{green}{with} each fault-triggering test. This allowed us to better understand how \textcolor{green}{the} fault was identified and resolved over time with respect to the changes in \textcolor{green}{fault-triggering} tests. \textbackslash peter\{maybe remove this, I don't quite get this detail\}Note that if a test is inherited from a parent class, we perform our analysis directly on \textcolor{green}{the} parent class since any changes would only occur in \textcolor{green}{that} class. & no \\
\hline
\end{tabularx}
\begin{center}
    \small \textbf{Table 4:} Sample of annotated paragraph pairs, identical chain of texts are in green.
    \label{annex:examples}
\end{center}


\end{document}